\documentclass{article}
\usepackage{bbm}

\usepackage{graphicx} 
\usepackage[toc,page]{appendix}
\usepackage{float}
\usepackage{hyperref}
\usepackage{courier}
\usepackage{natbib}
\usepackage{fancyhdr}
\usepackage{courier}
\usepackage{url}
\usepackage{multirow}
\usepackage{adjustbox}
\usepackage{amsmath}
\usepackage{amssymb}
\usepackage[inkscapeformat=png]{svg}
\usepackage{svg}
\usepackage{algorithm}
\usepackage{algorithmic}
\usepackage{graphicx}

\usepackage{etoolbox}

    \usepackage[preprint]{neurips_2023}

\usepackage[utf8]{inputenc} 
\usepackage[T1]{fontenc}    
\usepackage{hyperref}       
\usepackage{url}            
\usepackage{booktabs}       
\usepackage{amsfonts}       
\usepackage{nicefrac}       
\usepackage{microtype}      
\usepackage{xcolor}         

\title{DeepChem Equivariant: SE(3)-Equivariant Support in an Open-Source Molecular Machine Learning Library}

\author{
Jose Siguenza$^1$ \\
$^1$Deep Forest Sciences \\
\texttt{jose@deepforestsci.com}
 \And
  Bharath Ramsundar$^1$\\
  $^1$Deep Forest Sciences \\
  \texttt{bharath@deepforestsci.com} \\
}

\begin{document}

\maketitle

\begin{abstract}
Neural networks that incorporate geometric relationships respecting SE(3) group transformations (e.g. rotations and translations) are increasingly important in molecular applications, such as molecular property prediction, protein structure modeling, and materials design. These models, known as SE(3)-equivariant neural networks, ensure outputs transform predictably with input coordinate changes by explicitly encoding spatial atomic positions. Although libraries such as \textsc{e3nn} \citep{geiger2022e3nneuclideanneuralnetworks} and \textsc{SE(3)-Transformer} \cite{fuchs2020se3transformers} offer powerful implementations, they often require substantial deep learning or mathematical prior knowledge and lack complete training pipelines. We extend \textsc{DeepChem} \cite{Ramsundar-et-al-2019} with support for ready-to-use equivariant models, enabling scientists with minimal deep learning background to build, train, and evaluate models, such as SE(3)-Transformer and Tensor Field Networks. Our implementation includes equivariant models, complete training pipelines, and a toolkit of equivariant utilities, supported with comprehensive tests and documentation, to facilitate both application and further development of SE(3)-equivariant models.
\end{abstract}

\section{Introduction}
\label{submission}

Deep learning (DL) architectures such as Graph Neural Networks (GNNs) \cite{Veli_kovi__2023} and Transformers \cite{NIPS2017_3f5ee243} have driven significant progress in regression and classification tasks. These models have found wide application in several domains, including molecular property prediction \cite{le2022equivariantgraphattentionnetworks}, computer graphics \cite{Krzywda_2022}, robotics \cite{lin2022efficientinterpretablerobotmanipulation}, and materials science \cite{lee2025castcrossattentionbased}. Capturing spatial relationships is essential to improve model generalization and predictive accuracy in domains where structural and geometric information is fundamental. Equivariant neural networks incorporate geometric priors that ensure that predictions remain consistent under spatial changes such as rotations and translations, as defined by the SE(3) group \cite{satorras2022enequivariantgraphneural}. This is particularly important in molecular DL, where scalar properties must remain invariant and vector quantities, such as forces and dipole moments, must transform equivariantly \cite{liao2023equiformerequivariantgraphattention}. Equivariant models have enabled breakthroughs in drug discovery by improving the accuracy of molecular property prediction, protein–ligand binding affinity estimation, and molecular conformation generation \cite{schneuing2024structurebaseddrugdesignequivariant, nguyen2025equicpise3equivariantgeometricdeep}.

 Multiple SE(3)-equivariant models have been open-sourced, marking significant progress in symmetry-aware modeling. These implementations focus mainly on the model architecture and lack integration with broader machine learning workflows. For instance, \textsc{e3nn} \citep{geiger2022e3nneuclideanneuralnetworks} is a leading open-source library that provides a reliable and well-maintained infrastructure for equivariant modeling. However, it has limitations in dataset integration, data pre-processing, training pipelines, and readily available model architectures. Furthermore, mathematical depth and complexity present significant barriers to adoption, especially for researchers without DL and math expertise. In addition, repositories  like \textsc{SE(3)-Transformer} \cite{fuchs2020se3transformers}, \textsc{Cormorant} \cite{anderson2019cormorantcovariantmolecularneural}, offer equivariant models but pose challenges from deprecated dependencies, limited maintenance, and lack of continuous integration (CI), making them difficult to use in practical molecular DL workflows.

We introduce SE(3)-equivariant modeling capabilities to DeepChem \cite{Ramsundar-et-al-2019} by integrating equivariant models, utilities, and molecular featurization. This modular infrastructure broadens the application of SE(3)-equivariant methods, supporting models like SE(3)-Transformers and Tensor Field Networks (TFNs), and provides an accessible toolkit for building models that respect 3D geometric symmetries. Our framework efficiently computes equivariant features using spherical harmonics and irreducible representations, offering a practical and well-documented suite of tools for equivariant molecular modeling. The integrated equivariant framework is continuously tested and maintained within DeepChem, available at \url{https://github.com/deepchem/deepchem}, ensuring robustness and fostering contributions from our community.

\section{Background and Related Work}

We focus on learning from molecular data represented as 3D graphs, where spatial geometry and chemical context are relevant to accurate property prediction. Each molecule is modeled as a graph with atoms as nodes, defined by 3D coordinates \( x_i \in \mathbb{R}^3 \) and atomic features such as element type. Edges encode bond type and distance-based features.

\subsection{Equivariance and SE(3) symmetry}
\label{def:eqv}Let a group \( G \) act on two sets \( X \) and \( Y \). A function \( f: X \to Y \) is said to be equivariant with respect to the group action if it commutes with the action of every group element. Formally, this is defined as:

\begin{align}
f(g \cdot x) = g \cdot f(x), \quad \forall g \in G,\; x \in X
\end{align}

This means that applying a group transformation \( g \) to the input \( x \) before \( f \) yields the same result as applying \( f \) first and then transforming the output by \( g \). The function \( f \) is compatible with the action of \( G \). 

Computing an angular basis is essential for characterizing how tensors of varying ranks transform under these symmetries. This is achieved using irreducible representations, non-decomposable linear representations of the SE(3) group and spherical harmonics \cite{unke2024e3xmathrme3equivariantdeeplearning}. Irreducible representations are obtained via the Wigner D-matrix, which is closely related to the Clebsch–Gordan coefficients used in the coupling of angular momenta (see Appendix \ref{sec:cg-decomposition}). Denoted as $Y_{\ell}^{m}(\hat{\mathbf{r}})$, spherical harmonics provide a continuous angular basis aligned with SO(3) operations and encode information in the angular direction. These functions are defined in the unit sphere depending on the degree of angular momentum $\ell$ and order $m$. Spherical harmonics are combined with learnable radial functions to construct rotation-equivariant kernels and are described as follows:

\begin{align}
\label{eq2}
\Phi_{\ell}^{(i \to j)}(\mathbf{r}_{ij}) &= R_\ell(\|\mathbf{r}_{ij}\|) \, Y_{\ell}^{m}(\hat{\mathbf{r}}_{ij})
\end{align}

where $\mathbf{r}_{ij}$ is the relative distance between nodes $i$ and $j$, $R_\ell$ is a learnable radial function, and $Y_{\ell}^{m}$ captures angular dependences. Eq. \ref{eq2} defines the basis function $\Phi_{\ell}^{(i \to j)}(\mathbf{r}_{ij})$ in terms of its radial part $R_\ell(\|\mathbf{r}_{ij}\|)$ and its angular part, derived from spherical harmonics $Y_{\ell}^{m}(\hat{\mathbf{r}}_{ij})$.

In the context of SE(3)-equivariant networks, the equivariant weight basis (denoted as $\mathbf{K}_{\text{filter}}^{lk}$ or $\mathbf{K}_J$ in code implementation) is constructed to couple the input features (degree $k$), the output features (degree $l$), and the angular information. Equivariant weight basis ($\mathbf{K}_{\text{filter}}^{lk}$), incorporating the $\mathbf{Q}_J^{lk}$ matrix is constructed as:

\begin{align}
\label{eq4}
\mathbf{K}_{\text{filter}}^{lk}(\mathbf{r}_{ij}) &= R_{l k J}(\|\mathbf{r}_{ij}\|) \cdot \left( \mathbf{Q}_J^{lk} \cdot \mathbf{Y}_{J}(\hat{\mathbf{r}}_{ij}) \right)^T
\end{align}
This formulation ensures that equivariant feature vectors are computed and messages passed between nodes transforms equivariantly under $\mathrm{SE}(3)$ operations. Details about  kernel constraints and construction, Eq. \ref{eq4}, are presented in Appendix \ref{sec:cg-decomposition}, Eq. \ref{eq25} and Eq. \ref{eq26}) 
\subsection{Attention mechanism}
Attention mechanisms are widely used in DL to model structured data such as sequences, text, and molecular graphs. They enable context-aware weighting of input features, allowing models to capture complex relationships present in the input data.

\label{def:attn}
Consider a set of query vectors \( \mathbf{q}_i \in \mathbb{R}^d \) for \( i = 1, \ldots, m \), key vectors \( \mathbf{k}_j \in \mathbb{R}^d \) and value vectors \( \mathbf{v}_j \in \mathbb{R}^{d'} \) for \( j = 1, \ldots, n \), where \( d \) and \( d' \) denote the dimensions of the features. Each query vector attends to the value vectors by computing weights based on similarity with keys. The output of attention for a query \( \mathbf{q}_i \) is defined as:

\begin{align}
&\text{Attention}(\mathbf{q}_i, \{\mathbf{k}_j\}, \{\mathbf{v}_j\}) = \sum_{j=1}^{n} \alpha_{ij} \mathbf{v}_j,
\quad 
\end{align}
\begin{align}                             \alpha_{ij} = \frac{\exp(\mathbf{q}_i^\top \mathbf{k}_j)}{\sum_{j'=1}^{n} \exp(\mathbf{q}_i^\top \mathbf{k}_{j'})},
\end{align}

where the coefficients \( \alpha_{ij} \) are computed via a softmax function, ensuring they form a normalized distribution over the keys for each query. 

In \textsc{SE(3)-Transformer} \cite{fuchs2020se3transformers} and in our work, the standard self attention mechanism, which is central to the Transformer architecture \cite{vaswani2023attentionneed}, is adapted to respect SE(3) symmetry. This adaptation involves projecting feature spaces onto specific geometric representations before computing attention. The resulting self attention computes attention weights that are invariant and value embeddings that are equivariant under three dimensional roto translations. These properties ensure consistent predictions under spatial transformations while enabling efficient processing of molecular graphs (see Appendix~\ref{sec:slf-att}).

\section{Method}
\begin{figure}[h!]
\vskip 0.2in
\begin{center}
\centerline{\includegraphics[width=\columnwidth]{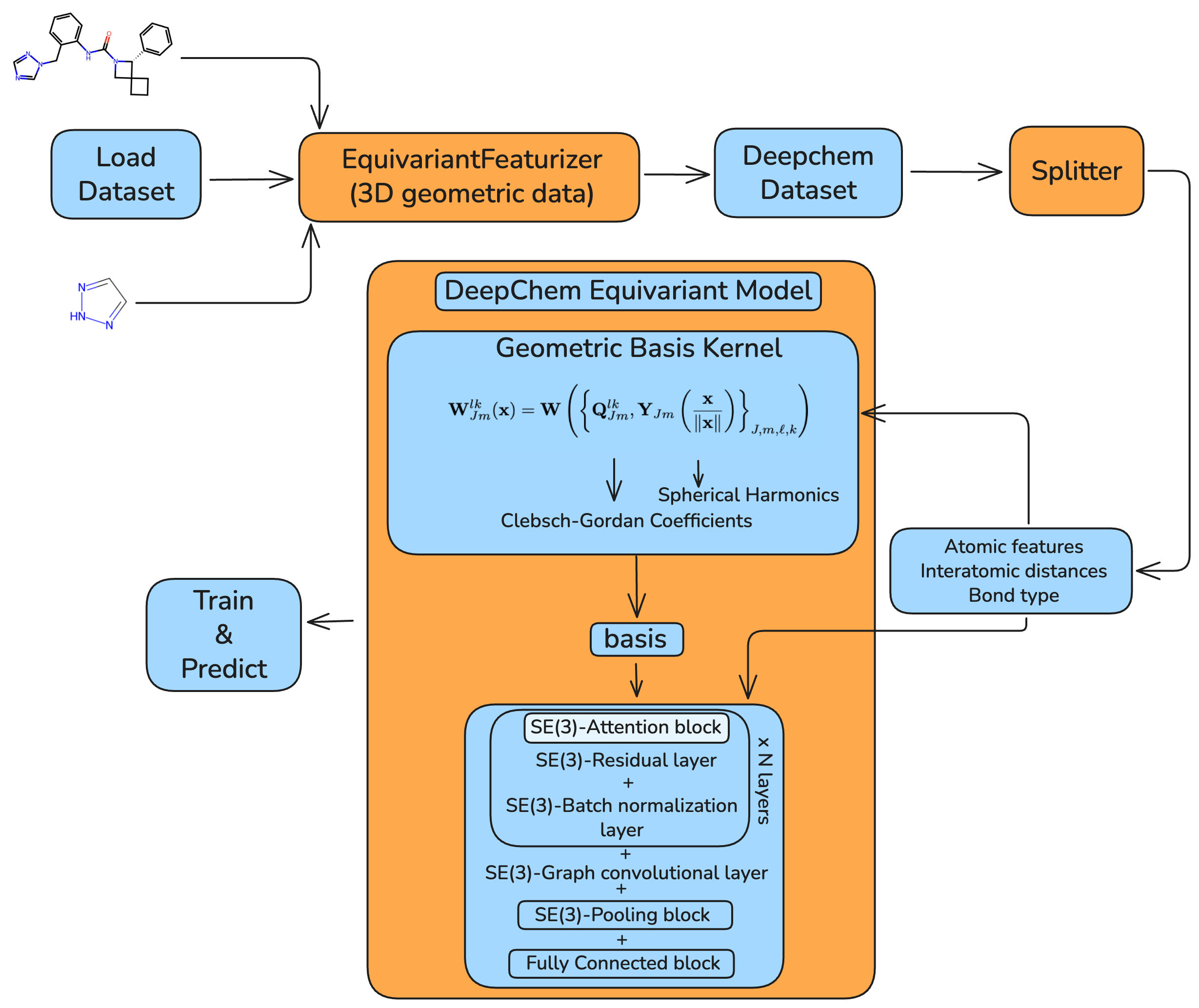}}
\caption{DeepChem pipeline for SE(3)-Transformer model. Input molecular data undergoes 3D geometric featurization, dataset preparation, and splitting, followed by equivariant model implementation and training using DeepChem's API.}
\label{eqv-pipeline}
\end{center}
\vskip -0.2in
\end{figure}

An overview of the main workflow enabling SE(3)-equivariant model support in DeepChem is presented in Figure \ref{eqv-pipeline}. This pipeline illustrates key components involved in data preprocessing, model construction, and training.
\subsection{SE(3)-Transformers}
 In our problem, SE(3)-Transformers operates on molecular graphs \( G = (V, E) \), where each node \( v_i \in V \) represents an atom with associated scalar features \( x_i \in \mathbb{R}^F \) and 3D coordinates \( p_i \in \mathbb{R}^3 \), and each edge \( e_{ij} \in E \) encodes pairwise features such as interatomic distances. Our implementation is inspired by the design and methodology of \textsc{SE(3)-Transformer} \cite{fuchs2020se3transformers} work.

As shown in Algorithm~\ref{alg:se3transformer}, input features are transformed into SE(3)-equivariant representations using spherical harmonics and tensor products. Computed equivariant features then pass through \(N\) \texttt{SE3ResidualAttention} layers applying multi-head equivariant attention, each followed by \texttt{SE3GraphNorm} for training stability. The weighted equivariant features are refined by \texttt{SE3GraphConv}, an SE(3)-equivariant graph convolution performing message passing with learned radial functions and angular kernels. Global pooling, via \texttt{SE3MaxPooling} or \texttt{SE3AvgPooling}, aggregates features into a graph-level embedding, which fully connected layers then map to the final output \(\hat{y} \in \mathbb{R}^T\), where \(T\) is the number of prediction tasks.

\begin{minipage}{0.7\linewidth}
\begin{algorithm}[H]
   \caption{SE(3)-Transformer}
   \label{alg:se3transformer}
\begin{algorithmic}
  \STATE \textbf{Input:} Graph $G$ with node features $x_i$ and 3D positions $p_i$
   \STATE \textbf{Parameters:} degrees $L$, channels $C$, layers $N_{\text{layers}}$

   \STATE $(\textit{basis}, r) \gets \texttt{get\_equivariant\_basis}(G, L-1)$
   \STATE $h \gets x_i$ \texttt{/* degree-0 features */}
   \FOR{$i = 1$ TO $N_{\text{layers}}$}
       \STATE $h \gets \texttt{SE3ResidualAttention}(h, G, r, \text{basis})$
       \STATE $h \gets \texttt{SE3GraphNorm}(h)$
   \ENDFOR
   \STATE $h \gets \texttt{SE3GraphConv}(h, G, r, \text{basis})$
   \STATE $h \gets \texttt{SE3Pooling}(h, \text{mode})$
   \texttt{/* mode: avg or max */}
   \STATE $h \gets \texttt{FullyConnected}(h)$
   \STATE \textbf{Return:} $h$
\end{algorithmic}
\end{algorithm}
\end{minipage}

\subsection{Residual equivariant attention}

Given a molecular graph \( G = (V, E) \) and node features \( h_i^{(d)} \) of degree \( d \), the layer computes attention in three stages: projection, weighting, and aggregation. First, query (\( \mathbf{q} \)), key (\( \mathbf{k} \)), and value (\( \mathbf{v} \)) tensors are constructed from input features using equivariant projection layers. Specifically, keys (\( \mathbf{k}_{ij} \)) and values (\( \mathbf{v}_{ij} \)) features are obtained via edge-based SE(3) convolutions (via \texttt{SE3PartialEdgeConv} layer), while queries (\( \mathbf{q}_i \)) features are produced through self-interaction (\texttt{SE3SelfInteraction} layer):
\begin{align}
\label{eq7}
\mathbf{q}_i = \mathcal{W}^{(q)} h_i, \quad
\mathbf{k}_{ij} = \mathcal{W}^{(k)} h_j, \quad
\mathbf{v}_{ij} = \mathcal{W}^{(v)} h_j.
\end{align}

In Eq. \ref{eq7}, $\mathcal{W}^{(i)}$, where $i \in {q, k, v}$, denotes the equivariant weight basis defined as in Eq. \ref{eq4}. In our DeepChem-Equivariant implementation, this basis is constructed within the \texttt{SE3PairwiseConv} layer.

Next, attention weights are computed using a generalized dot-product across tensor features:
\begin{align}
\alpha_{ij} = \frac{\langle \mathbf{q}_i, \mathbf{k}_{ij} \rangle}{\sqrt{d}} \quad 
\end{align}
where \( \langle \cdot , \cdot \rangle \) represents the inner product that respects the tensor structure.

The final attention-weighted output is computed as:
\begin{align}
z_i = \sum_{j \in \mathcal{N}(i)} \alpha_{ij} \cdot \mathbf{v}_{ij}.
\end{align}

To ensure effective training, the layer includes a skip connection mechanism. If the skip type is 'sum', a learnable self-interaction \( \mathcal{W}^{\text{self}} h_i \) is added to the output:
\begin{align}
h_i^{\prime} = z_i + \mathcal{W}^{\text{self}} h_i.
\end{align}
If 'cat' is used, the input and output tensors are concatenated and projected back to the output space.

\begin{minipage}{0.7\linewidth}
\begin{algorithm}[H]
\caption{SE(3)-Equivariant Residual Attention}
\label{alg:se3residualattention}
\begin{algorithmic}
\STATE \textbf{Input:} Graph $G$ with features $\mathbf{h}$, distances $r$, equivariant basis
\STATE \textbf{parameters:} \#Heads $H$, skip type $\in \{\text{sum}, \text{cat}, \text{None}\}$

\STATE $q \gets \texttt{SE3SelfInteraction}$

\STATE $k, \textit{v} \gets \texttt{SE3PartialEdgeConv}(\mathbf{h}, G, r, \text{basis})$
\STATE $z \gets \texttt{SE3MultiHeadAttention}(\mathbf{q}, \mathbf{k}, \mathbf{v}, G)$

\IF{skip = cat}
    \STATE $z \gets \texttt{SE3SelfInteraction}(\texttt{SE3Cat}(\mathbf{z}, \mathbf{h}))$
\ELSIF{skip = sum}
    \STATE $z \gets \texttt{SE3Sum}(\texttt{SE3SelfInteraction}(\mathbf{z}), \mathbf{h})$
\ENDIF

\STATE \textbf{Return:} $\mathbf{z}$ \texttt{/* Updated features */}

\end{algorithmic}
\end{algorithm}
\end{minipage}
\subsubsection{Localized attention on graphs}
\label{local-gph-att}
Applying self-attention directly to molecular graphs scales quadratically with the number of nodes, making it computationally expensive for large inputs. To address this, attention is restricted to local neighborhoods, each node attends only to a fixed set of nearby nodes. A general form of localized attention over graph neighborhoods $\mathcal{N}_i$ is presented in \cite{fuchs2020se3transformers}:

\begin{align}
\mathbf{y}_i = \frac{1}{C(\{ \mathbf{f}_j \in \mathcal{N}_i \})} \sum_{j \in \mathcal{N}_i} w(\mathbf{f}_i, \mathbf{f}_j) \cdot h(\mathbf{f}_j),
\end{align}

where $\mathbf{f}_i$ and $\mathbf{f}_j$ are node features, $w$ and $h$ are learnable functions, and $C$ is a normalization factor dependent on the neighborhood.

\subsubsection{Self-attention mechanism and permutation invariance}
\label{sec:slf-att}
In the special case of self-attention, the query, key, and value vectors are all derived from the same set of input features \( \mathbf{f}_i \in \mathbb{R}^d \). Specifically, we define:

\begin{align}
\mathbf{q}_i = W_Q \mathbf{f}_i, \quad 
\mathbf{k}_i = W_K \mathbf{f}_i, \quad 
\mathbf{v}_i = W_V \mathbf{f}_i,
\end{align}

where \( W_Q, W_K, W_V \) are learnable projection matrices. Each element attends to other elements in the set, enabling contextual representation learning.

A key property of attention is permutation equivariance. When applied to a set of inputs, such as graph nodes, the output of attention remains consistent under any permutation of the inputs. This makes attention particularly suitable for unordered data like 3D point clouds or molecular graphs, where geometric relationships dictate how the object structure is composed.

The SE(3)-equivariant residual attention layer, shown in Algorithm~\ref{alg:se3residualattention}, extends the attention mechanism to 3D molecular graphs while preserving equivariance under SE(3) group operations. The node features are projected into queries via \texttt{SE3SelfInteraction}, and keys/values are computed using \texttt{SE3PartialEdgeConv}, incorporating edge features, pairwise distances $r$, and SE(3)-equivariant basis functions. Attention is applied through \texttt{SE3MultiHeadAttention}, aggregating information across graph edges.

The output of the attention mechanism is then added back to the original input features, forming a residual connection (using \texttt{SE3Sum}). Alternatively, it can be concatenated with the input (\texttt{SE3Cat}). This operation yields an updated tensor field $\mathbf{z}$ that supports geometry-aware message passing.

\subsection{SE(3) equivariant graph convolutions}
SE(3)-equivariant graph convolution layer, presented in Algorithm \ref{alg:se3graphconv}, propagates information across molecular graphs considering roto-translation constraints. Each node \( v_i \) aggregates messages from its neighbors \( \mathcal{N}(i) \) based on both scalar and tensor features, using kernels constructed from spherical harmonics and learnable radial functions.

\begin{minipage}{0.7\linewidth}
\begin{algorithm}[H]
\caption{SE(3)-Equivariant Graph Convolution}
\label{alg:se3graphconv}
\begin{algorithmic}
\STATE \textbf{Input:} Graph $G$, node features $h$, distances $r$, basis tensors $basis$
\STATE \textbf{parameters:} self-interaction flag, convolution flavor $\in \{\text{TFN}, \text{skip}\}$

\FOR{degree $d$ in $h$} 
    \STATE $G.ndata[d] \gets h[d]$
\ENDFOR

\STATE \textbf{/* Build edge features */}
\STATE $feat \gets$ concat$(w, r)$ if edge weights $w$ exist else $r$

\STATE \textbf{/* Compute kernels */}
\FOR{$(d_{in}, d_{out})$}
    \STATE $G.\textit{edata}[(d_{\text{in}}, d_{\text{out}})] \gets \texttt{SE3PairwiseConv}(\text{feat}, \text{basis})$

\ENDFOR

\STATE \textbf{/* Message passing */}
\FOR{ $d_{out}$}
    \STATE $G.\textit{update\_all}(\textit{edge\_fn}(d_{\text{out}}),\ \textit{mean} \rightarrow d_{\text{out}})$

\ENDFOR

\STATE \textbf{Return:} $\mathbf{h}' \leftarrow \{G.\textit{ndata}[d]\}$ 

\texttt{/* Dictionary of updated tensors for each degree $d$ */}
\end{algorithmic}
\end{algorithm}
\end{minipage}

For every pair of input and output degrees \( (d_{\text{in}}, d_{\text{out}}) \), \texttt{SE3GraphConv} layer computes convolutional kernels using the angular of SE(3) and the interatomic distances, applying \texttt{SE3PairwiseConv} layer. In \texttt{SE3PairwiseConv}, the radial component is included using \texttt{SE3RadialFunc}. These kernels are applied to incoming messages through pairwise convolutions, which are accumulated across edges. Optionally, a self-interaction term is added allowing nodes to preserve their original features or perform residual updates.

\subsubsection{Equivariant message passing and TFNs}

Output features follows TFN style or skip-connections and are computed for each node and degree \( d \) as:
\begin{align}
h^{\prime(d)}_i = \text{Mean}_{j \in \mathcal{N}(i)} \left[ \mathcal{K}_{ij}^{(d)} * h_j^{(d_{\text{in}})} \right] + \delta_{d_{\text{in}}, d} \cdot \mathcal{W}^{\text{self}} h_i^{(d)},
\end{align}
where \( \mathcal{K}_{ij}^{(d)} \) denotes the kernel matrix and \( \delta \) is the Kronecker delta selecting self-interaction only for matching degrees. \( \mathcal{W}^{\text{self}} h_i^{(d)} \) is a learnable self-interaction transformation, analogous to a residual connection, allowing node \( i \) to retain and adapt its own feature representation.

This message-passing operation enables the layer to learn localized, symmetry-aware representations of graphs that are sensitive to spatial arrangement and feature type.

\textbf{Tensor Field Networks} TFNs are SE(3)-equivariant architectures designed for learning over meshes, point clouds, and graphs \cite{thomas2018tensorfieldnetworksrotation}. Given a set of 3D points \( \{\mathbf{x}_j\} \) with associated features \( \{\mathbf{f}_j\} \), TFNs interpret the inputs as vector fields and apply convolution with rotation-equivariant kernels. Each TFN layer maps input features of type-\( k \) to output features of type-\( \ell \) using a learnable kernel \( W^{\ell k}(\mathbf{x}) \), presented in Eq. \ref{eq4}, which is decomposed into a sum over basis kernels constrained by symmetry.

Message passing in TFNs is performed by aggregating these equivariant convolutions across local neighborhoods. To ensure expressive updates, a self-interaction term is added when \( \ell = k \), enabling transformations via learned scalar weights.

\begin{minipage}{0.7\linewidth}
\begin{algorithm}[H]
   \caption{TFN: Tensor Field Network}
   \label{alg:tfn}
\begin{algorithmic}
   \STATE \textbf{Input:} Graph $G$ with node features $x_i$ and 3D positions $p_i$
   \STATE \textbf{Parameters:} degrees $L$, channels $C$, layers $N_{\text{layers}}$
   \STATE $(\textit{basis}, r) \gets \texttt{get\_equivariant\_basis}(G, L-1)$
   \STATE $h \gets x_i$ \texttt{/* degree-0 features */}
   \FOR{$i = 1$ TO $N_{\text{layers}}$}
       \STATE $h \gets \texttt{SE3GraphConv}(h, G, r, \text{basis})$
       \STATE $h \gets \texttt{SE3GraphNorm}(h)$
   \ENDFOR
   \STATE $h \gets \texttt{SE3GraphConv}(h, G, r, \text{basis})$
   \STATE $h \gets \texttt{SE3Pooling}(h, \text{mode})$ \texttt{/* avg or max */}
   \STATE $h \gets \texttt{FullyConnected}(h)$
   \STATE \textbf{Return:} $h$
\end{algorithmic}
\end{algorithm}
\end{minipage}

\subsection{DeepChem Integration}

We integrated the core components for SE(3)-equivariant modeling into DeepChem's ecosystem. This involved adapting layers of the SE(3)-Transformer architecture \cite{fuchs2020se3transformers}, including spherical harmonics computations, to a modular and extensible API (see Appendix \ref{tab:sh-core-components}). To support structured molecular input, we implemented the \texttt{EquivariantGraphFeaturizer}, which transforms QM9 molecules from MoleculeNet into rich graph-based representations that capture both geometric and chemical information. To ensure correctness and reproducibility, we developed unit tests to verify SE(3) equivariance and CI workflows using GitHub actions.

\section{Experiments}

We used the QM9 dataset from MoleculeNet \cite{wu2018moleculenetbenchmarkmolecularmachine} to evaluate Deepchem's SE(3)-Transformer and TFN implementations for molecular property prediction. A total of 132,480 molecules were included, with the data set divided into training (80\%), validation (10\%), and test (10\%) sets. Molecules in QM9 consist of up to 29 atoms and five atomic species: carbon, nitrogen, oxygen, hydrogen, and fluorine. Interatomic distances are measured in ångtröms (Å). We applied the same model architecture and hyperparameters as in \cite{fuchs2020se3transformers}, which were optimized for the regression task to predict $\varepsilon_{\mathrm{HOMO}}$. For detailed model training specifications, see Appendix \ref{sec:qm9-training-details}. Table \ref{tab:qm9results} shows the mean absolute error (MAE) results for molecular properties reported in \cite{fuchs2020se3transformers}, where lower values indicate better performance. Our method achieves comparable results to Cormorant\cite{anderson2019cormorantcovariantmolecularneural}, TFN \cite{thomas2018tensorfieldnetworksrotation}, and SE(3)-Transformer \cite{fuchs2020se3transformers}, from which our implementation is adapted using irreducible SE(3) representations.  \textsc{DeepChem-Equivariant} relies on computationally expensive higher-order representations, while, \textsc{LieConv} offers an efficient alternative by learning equivariant filters directly using Lie algebra. We plan to implement \textsc{LieConv} in \textsc{DeepChem-Equivariant} in future work.

\begin{table}[h!]
    \centering
    \caption{{Mean Absolute Error on the QM9 regression task for isotropic polarizability ($\alpha$), HOMO energy ($\epsilon_{\text{HOMO}}$), LUMO energy ($\epsilon_{\text{LUMO}}$), HOMO-LUMO gap ($\Delta\epsilon$), dipole moment ($\mu$), and heat capacity at 298.15K ($C_v$). Results are reported as the mean and standard deviation of triplicate measurements.}}
    \resizebox{\linewidth}{!}{
    \begin{tabular}{|l|c|c|c|c|c|c|}
        \hline
        \textbf{Model} & $\alpha$ [bohr$^3$]& $\Delta\epsilon [meV]$ & $\epsilon_{\text{HOMO}}$ [meV] & $\epsilon_{\text{LUMO}}$ [meV] & $\mu$ [D]& $C_v$ [kcal/mol]\\
        \hline
        \multicolumn{7}{|c|}{\textbf{Non-Equivariant Models}} \\
        \hline
        WaveScatt \cite{Hirn_2017} & .160 & 118 & 85 & 76 & .340 & .049 \\
        NMP \cite{gilmer2017neuralmessagepassingquantum} & .092 & 69 & 43 & 38 & \textbf{.030} & .040 \\
        SchNet \cite{schütt2017schnetcontinuousfilterconvolutionalneural} & .235 & 63 & 41 & 34 & .033 & .033 \\
        \hline
        \multicolumn{7}{|c|}{\textbf{Equivariant Models}} \\
        \hline
        Cormorant \cite{anderson2019cormorantcovariantmolecularneural} & .085 & 61 & 34 & 38 & .038 & \textbf{.026} \\
        LieConv (T3) \cite{pmlr-v119-finzi20a} & \textbf{.084} & \textbf{49} & \textbf{30} & \textbf{25} & .032 & .038 \\
        TFN \cite{thomas2018tensorfieldnetworksrotation} & .223 & 58 & 40 & 38 & .064 & .101 \\
        SE(3)-Transformer \cite{fuchs2020se3transformers} & .142 & 53 & 35 & 33 & .051 & .054 \\
        \hline
        \multicolumn{7}{|c|}{\textbf{DeepChem-Equivariant}} \\
        \hline
        \textbf{SE(3)-Transformer} & .182 $\pm$ .006 & 62 $\pm$ 0.9 & 39 $\pm$ 1.1 & 38 $\pm$ .3 & .071 $\pm$ .005 & .089 $\pm$ .003 \\
        \textbf{TFN} & .192 $\pm$ .008 & 64 $\pm$ 2.9 & 38 $\pm$ .4 & 39 $\pm$ .3 & .065 $\pm$ .006 & .086 $\pm$ .004 \\
        \hline
    \end{tabular}}
    \label{tab:qm9results}
\end{table}
\section{Conclusion}

We introduce SE(3)-equivariant support to DeepChem, comprising utility functions, data processing tools, and a comprehensive training pipeline tailored for molecular graphs. The SE(3)-Transformer model developed with this toolkit maintains invariance to rotations and translations, eliminating the need for data augmentation and enhancing robustness across coordinate systems. Compared with established frameworks such as \textsc{e3nn} \cite{geiger2022e3nneuclideanneuralnetworks} and \textsc{SE(3)-Transformer} \cite{fuchs2020se3transformers}, our approach prioritizes usability, dataset integration, and long-term maintenance, expanding access to equivariant modeling in molecular DL. Experiments indicated that incorporating attention into roto-translation-equivariant models consistently improves accuracy and training stability. While performance could improve compared with other equivariant baselines, efficiency remains a challenge, particularly in basis construction. Finally, caching precomputed bases in DeepChem’s \texttt{GraphData} class is a promising enhancement to avoid bottlenecks during training, improve scalability, and promote open-source collaboration.

\section*{Impact Statement}

This work presents DeepChem's support of symmetry-aware machine learning models, with a focus on applications in molecular modeling, drug discovery, and materials science. By integrating SE(3)-equivariant architectures, we aim to lower the barrier for researchers to apply geometric DL methods.


\bibliographystyle{abbrv}

\newpage
\appendix
\onecolumn
\section{Computation of Equivariant Features}

\subsection{Spherical Harmonics:}

\begin{align}
Y_{Jm}(\theta, \phi) = \sqrt{\frac{2J + 1}{4\pi} \cdot \frac{(J - |m|)!}{(J + |m|)!}} \cdot P_J^{|m|}(\cos \theta) \cdot
\begin{cases}
\sin(|m|\phi), & m < 0, \\
1, & m = 0, \\
\cos(m\phi), & m > 0.
\end{cases}
\end{align}

\paragraph{Legendre Polynomial Definitions and Properties:}

\begin{align}
P_J^{|J|}(x) &= (-1)^{|J|} (1 - x^2)^{|J|/2} (2|J| - 1)!! 
\quad &&\text{(Boundary: } J = m\text{)}
\end{align}

\begin{align}
P_J^{-m}(x) &= (-1)^m \frac{(J - |m|)!}{(J + |m|)!} P_J^{|m|}(x) 
\quad &&\text{(Negative } m \text{)}
\end{align}
\begin{align}
P_J^{|m|}(x) &= \frac{2J - 1}{J - |m|} x P_{J-1}^{|m|}(x) + \mathbbm{1}[J - |m| > 1] \cdot \frac{J + |m| - 1}{J - |m|} P_{J-2}^{|m|}(x) 
\quad &&\text{(Recursion)}
\end{align}
Our implementation, adapted from \cite{fuchs2020se3transformers}, defines real-valued spherical harmonics \(Y_{Jm}(\theta, \phi)\), incorporating normalization, associated Legendre polynomials, and angular components.  
The formulation handles different \(m\) values using sine, cosine, or constant factors.  
Legendre polynomials \(P_J^m(x)\) are computed recursively and include special cases for boundary and negative orders.  
These functions are fundamental for expressing angular dependencies in equivariant models and spherical basis projections.

\subsection{Clebsch--Gordan Decomposition}
\label{sec:cg-decomposition}
In equivariant neural networks for 3D data, such as Tensor Field Networks and SE(3)-Transformers, equivariant kernels must satisfy specific transformation constraints under the action of SO(3) or SE(3). These constraints naturally involve the Clebsch--Gordan (CG) decomposition from the representation theory of compact Lie groups.

Let $D^{(l)}(g)$ denote the Wigner-D matrix representing an irreducible representation (irrep) of SO(3) with angular momentum $l$, and let $g \in \mathrm{SO}(3)$ be a rotation. The tensor product of two irreps $D^{(k)}(g)$ and $D^{(l)}(g)$ forms a new (generally reducible) representation:
\begin{align}
D^{(k)}(g) \otimes D^{(l)}(g)
\end{align}

This product can be decomposed into a direct sum of irreps:
\begin{align}
D^{(k)}(g) \otimes D^{(l)}(g) = Q_{kl}^\top \left( \bigoplus_{J=|k-l|}^{k+l} D^{(J)}(g) \right) Q_{kl}
\end{align}

where $Q_{kl}$ is the change-of-basis matrix composed of Clebsch--Gordan coefficients. These coefficients provide the transformation from the tensor product basis to the direct sum of irreducible representations. 

\paragraph{Kernel Constraint.}
In the context of convolutional operations on point clouds, we consider kernels $W_{lk}(\mathbf{x}) \in \mathbb{R}^{(2l+1) \times (2k+1)}$ that map type-$k$ features to type-$l$ features. The requirement of equivariance under SO(3) imposes the kernel constraint:

\begin{align}
W_{lk}(R^{-1}\mathbf{x}) = D^{(l)}(g) \, W_{lk}(\mathbf{x}) \, D^{(k)}(g)^{-1}, \quad \forall g \in \mathrm{SO}(3)
\end{align}

or, in vectorized form using the identity $\mathrm{vec}(AXB) = (B^\top \otimes A) \mathrm{vec}(X)$:

\begin{align}
\mathrm{vec}(W_{lk}(R^{-1}\mathbf{x})) = (D^{(k)}(g) \otimes D^{(l)}(g)) \, \mathrm{vec}(W_{lk}(\mathbf{x}))
\end{align}

Using the CG decomposition of the tensor product representation:

\begin{align}
D^{(k)}(g) \otimes D^{(l)}(g) = Q_{kl}^\top \left( \bigoplus_{J=|k-l|}^{k+l} D^{(J)}(g) \right) Q_{kl}
\end{align}

we define a new basis:

\begin{align}
\eta_{lk}(\mathbf{x}) := Q_{kl} \, \mathrm{vec}(W_{lk}(\mathbf{x})),
\end{align}

in which the kernel constraint becomes block-diagonal:

\begin{align}
\eta_{lk}(R^{-1}\mathbf{x}) = \left( \bigoplus_{J=|k-l|}^{k+l} D^{(J)}(g) \right) \, \eta_{lk}(\mathbf{x}).
\end{align}

This shows that each block of $\eta_{lk}$ transforms independently under a known irrep, and thus can be modeled using known basis functions such as spherical harmonics $Y_J(\mathbf{x})$:

\begin{align}
\label{eq25}
\eta_{lk}(\mathbf{x}) = \bigoplus_{J=|k-l|}^{k+l} c_J \, Y_J(\mathbf{x})
\end{align}

Transforming back give:

\begin{align}
\label{eq26}
\mathrm{vec}(W_{lk}(\mathbf{x})) = Q_{kl}^\top \, \eta_{lk}(\mathbf{x}),
\end{align}

which yields a complete solution to the equivariant kernel construction. The construction of equivariant kernels were dereived from \textsc{SE(3)-Transformer} \cite{fuchs2020se3transformers}, but we include our own implementation for irreducible representations including SU(2) generators and changes of basis from real to complex to get SO(3) generators used to compute Wigner-D matrices.

\subsection{DeepChem-Equivariant Utility Functions}
\label{tab:sh-core-components}
Table \ref{sh-table} summarizes core functions and classes for constructing spherical harmonics and equivariant transformations adapted from \cite{fuchs2020se3transformers}. These components facilitate angular basis transformations, rotational symmetries, and efficient spherical harmonic computations.

\begin{table}[t!]
\caption{Core functions and classes used to construct spherical harmonics and their equivariant transformations.The function \textit{basis\_transformation\_Q\_J} serves for computing equivariant angular momentum basis transformations. \texttt{SphericalHarmonics} is the primary utility for precomputing spherical harmonics used in constructing angular bases.}
\label{sh-table}
\vskip 0.15in
\begin{center}
\begin{small}
\begin{tabular}{ll}
\toprule
Component & Description \\
\midrule
\textit{basis\_transformation\_Q\_J} & Computes equivariant basis transformations for angular momentum \(J\). \\
\quad \textit{\_R\_tensor} & Calculates tensor products of rotation matrices for combined rotations. \\
\quad \textit{irr\_repr} & Generates irreducible SO(3) representations encoding rotational symmetries. \\
\qquad \textit{wigner\_D} & Computes Wigner D-matrices representing rotations in angular momentum basis. \\
\qquad\quad \textit{so3\_generators} & Constructs SO(3) Lie algebra generators for 3D rotation infinitesimals. \\
\qquad\quad\quad \textit{su2\_generators} & Builds SU(2) Lie algebra generators related to spin and SO(3) rotations. \\
\quad \textit{kron} & Performs Kronecker products to combine basis components in equivariant ops. \\
\quad \textit{matrices\_kernel} & Builds kernel matrices enforcing equivariance in neural network layers. \\
\quad \textit{\_sylvester\_submatrix} & Constructs Sylvester submatrices used in matrix transformations. \\
\midrule
\textit{SphericalHarmonics} & Class for computing and manipulating spherical harmonics values. \\
\quad \textit{lmpv} & Stores precomputed \((l, m, p)\) indices for efficient spherical harmonic lookup. \\
\quad \textit{pochammer} & Computes Pochhammer symbols for normalization in harmonic formulas. \\
\quad \textit{semifactorial} & Calculates semifactorials (double factorials) used in harmonic coefficients. \\
\midrule
\textit{get\_spherical\_from\_cartesian} & Converts Cartesian \((x,y,z)\) to spherical \((r, \theta, \phi)\) coordinates. \\
\bottomrule
\end{tabular}
\end{small}
\end{center}
\vskip -0.1in
\end{table}

\section{QM9 Training Details}
\label{sec:qm9-training-details}

Experiments were conducted using an NVIDIA A100-SXM4-40GB GPU, 4-core CPU, and 35 GB of RAM. We used a batch size of 72 and training took around 16.16 hours per molecular property.

We summarize the key hyperparameters used to train our SE(3)-Transformer model in Table~\ref{tab:training-hyperparams}. The model was configured with 7 equivariant layers, 32 channels, and 4 irreducible representation degrees, using a max pooling strategy and 8 attention heads. The input atom features were of size 6, and edge features of dimension 4 were used to capture pairwise interactions.

\begin{table}[b!]
\centering
\caption{SE(3)-Transformer Model Training Hyperparameters}
\label{tab:training-hyperparams}
\vskip 0.1in
\begin{small}
\begin{sc}
\begin{tabular}{ll}
\toprule
\textbf{Parameter} & \textbf{Value} \\
\midrule
Number of layers & 7 \\
Atom feature size & 6 \\
Number of workers & 4 \\
Number of channels & 32 \\
Number of nonlinearity layers & 0 \\
Number of degrees & 4 \\
Edge feature dimension & 4 \\
Pooling type & max \\
Number of attention heads & 8 \\
Batch size & 72 \\
Device & A100 (GPU) \\

\bottomrule
\end{tabular}
\end{sc}
\end{small}
\vskip -0.1in
\end{table}

\end{document}